\pdfoutput=1

\documentclass[11pt]{article}

\usepackage[preprint]{acl}

\usepackage{times}
\usepackage{latexsym}
\usepackage{amsmath}

\usepackage{amsfonts}
\usepackage{amssymb}
\usepackage{needspace}
\usepackage{xcolor}
\usepackage{framed}
\usepackage{enumitem}
\usepackage{comment}
\usepackage{makecell}
\usepackage{soul}

\usepackage{lettrine}

\usepackage[skins,breakable,listings]{tcolorbox}

\newtcblisting{promptbox}{
  breakable,
  colback=gray!5,
  colframe=gray!50!black,
  listing only,
  listing options={basicstyle=\ttfamily\small, breaklines=true}
}
  
\usepackage[T1]{fontenc}
\usepackage[utf8]{inputenc}
\usepackage{microtype}
\usepackage{booktabs}
\usepackage{inconsolata}
\usepackage{graphicx}
\usepackage{subcaption} 
\usepackage{enumitem}
\usepackage{multicol}
\usepackage{multirow}
\usepackage{diagbox}
\usepackage{fontawesome5}

\newcommand{\mathemagic}[0]{\textsc{MatheMagic}}

\title{\mathemagic{}: Generating Dynamic Mathematics \\Benchmarks Robust to Memorization}

\author{
  Dayyán O'Brien\textsuperscript{\raisebox{-0.3ex}{\includegraphics[width=2.1ex]{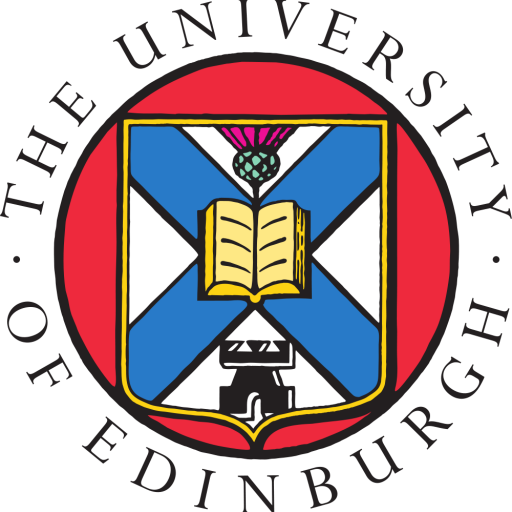}}} 
  \quad
  Barry Haddow\textsuperscript{\raisebox{-0.3ex}{\includegraphics[width=2.1ex]{logos/edinburgh.png}},\,\raisebox{-0.1ex}{\includegraphics[width=2ex]{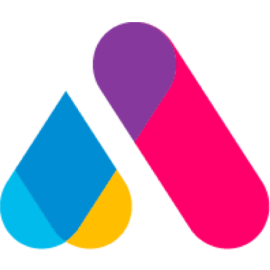}}}
  \quad
  Emily Allaway\textsuperscript{\raisebox{-0.3ex}{\includegraphics[width=2.1ex]{logos/edinburgh.png}}} 
  \quad
  Pinzhen Chen\textsuperscript{\raisebox{-0.3ex}{\includegraphics[width=2.1ex]{logos/edinburgh.png}},\,\raisebox{-0.1ex}{\includegraphics[width=2ex]{logos/aveni.png}},\,\raisebox{-0.05ex}{\includegraphics[width=1.6ex]{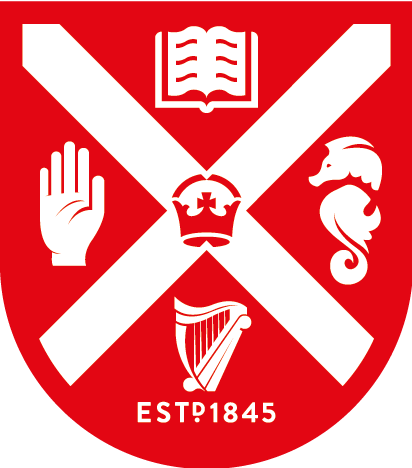}}}
  \\\\
  \raisebox{-0.3ex}{\includegraphics[width=2.1ex]{logos/edinburgh.png}}\hspace{0.2ex}University of Edinburgh
  \qquad
  \raisebox{-0.2ex}{\includegraphics[width=2ex]{logos/aveni.png}}\,Aveni
  \qquad
  \raisebox{-0.15ex}{\includegraphics[width=1.5ex]{logos/qub.png}}\,Queen's University Belfast
  \\
  \texttt{\{dayyan.obrien,emily.allaway\}@ed.ac.uk, barry@aveni.ai, p.chen@qub.ac.uk}
}

\begin{document}
\maketitle
\begin{abstract}
Conducting contamination-free evaluation of mathematical capabilities can be difficult for two reasons: models may memorize a test set once it is made public, and mathematical benchmarks are prone to overfitting due to having limited diversity of symbols and rules, coupled with closed-ended answers. We propose to leverage these shortcomings as useful features to construct a dynamic, counterfactual benchmark---to both reveal overfitting and measure reasoning over arithmetic. We demonstrate this via \mathemagic{}, which generates arithmetic test instances with the interpretations of numbers and operators altered, yet has automatically verifiable answers. Instances are randomly seeded and constructed at test time to evaluate a model's induction or deduction ability, offering stability, extensibility, comparability, and robustness to overfitting. Our experiments find that models solve deduction more easily than induction, but they revert to standard arithmetic. Further, math-adapted models are less flexible at following unfamiliar arithmetic rules, and fine-tuning on induction tasks generalizes poorly. \par\vspace{0.6ex}
\centering\faGithub\ \href{https://github.com/dayyanobrien/mathemagic}{\texttt{dayyanobrien/mathemagic}}  

\end{abstract}

\section{Introduction}
The main breakthrough of large language models (LLMs) is their ability to reason, allowing for complex planning and decision-making. Although reasoning is difficult to define or quantify, mathematical tasks have emerged as a popular testbed because of their verifiable answers, making evaluation straightforward. However, the closed‑ended nature allows models to rely on rote recall or pattern matching \citep[][inter alia]{mirzadeh2024gsm}, because math has relatively low variety in symbols and structure, making it susceptible to models memorizing combinations of math tokens. Furthermore, once a static test is publicized, it may leak into pre-training or post-training data, not to mention intentional efforts to overfit benchmarks \citep{sainz-etal-2023-nlp,wu2025reasoning}. Consequently, metric results can be inflated, leading to an overestimation of reasoning capability. Simple, static math benchmarks are therefore becoming less reliable. 

To tackle this, we propose a
methodology for constructing a dynamic and counterfactual benchmark. Our framework, \mathemagic{}, generates benchmarks designed to penalize models that have overfit to, or memorized, superficial math expression strings. To do this, we simply change the rules of arithmetic in grade-school arithmetic expressions. For instance, by remapping arithmetic symbols (e.g., defining the symbol \texttt{+} to mean multiplication) or applying operational changes to the entire expression (e.g., adding a fixed constant to each result). To correctly answer \mathemagic{} questions, a model must reason from its context only, providing a more reliable measure of inductive and deductive reasoning over arithmetic, as its pretrained arithmetic knowledge is (mostly) incorrect under these changes. A philosophy of our methodology is that although the questions are procedurally and dynamically generated, all answers remain verifiable, like in original mathematics. The process, controlled by a random seed, satisfies the duality of comparability and robustness to contamination. 

We evaluate the arithmetic reasoning of several LLMs using a benchmark generated by \mathemagic{}, which allows for customized rules and is scalable by design; for this study, we define 9 transformations and use 5 seeded test sets of 405 examples each. Through seven detailed analyses, we find that models perform well when a new rule is explained explicitly (deduction), but they struggle to infer from examples alone (induction). This failure evolves as models see more examples. Initially, the failure is rooted in a bias towards ingrained knowledge, causing reversions to standard arithmetic. As more examples are provided, this bias gives way to a struggle with procedural complexity, which is marked by a rise in novel errors as models attempt to apply the new rules. While we find that performance generally improves with more examples, it often hits a plateau or even degrades. Some of our transformations are harder than others, particularly those requiring multi-step reasoning, which prove far harder than a simple, operational change. This suggests the models' understanding of these rules is superficial, which we confirm with fine-tuning. 
Furthermore, this inductive ability fails to generalize through finetuning, leading models to overfit to the patterns they have seen. This leads to modest performance gains on tests in-domain and no meaningful improvement on most of the out-of-domain ones.

Our contributions can be outlined as follows:
\begin{enumerate}[leftmargin=3.5ex,noitemsep,topsep=0ex]
    \item A new method to benchmark arithmetic reasoning that is robust to memorization, through dynamic question generation, yet is accessible, reproducible, and easily extensible to new transformations.
    \item Seven detailed analyses, including different types of reasoning, memorization, and generalization in LLMs on arithmetic, as well as a fine-grained inspection of the tasks in the \mathemagic{} framework.
\end{enumerate}

\begin{table*}[t]
\centering\small
\resizebox{1\textwidth}{!}{%
\begin{tabular}{p{1.8cm}lp{8cm}p{6cm}}
\toprule
\textbf{Category} & \textbf{Transformation} & \textbf{Description} & \textbf{Example} \\
& & & (Base: \texttt{123 + 10 = 133}) \\
\midrule
Fundamental & \texttt{swap\_symbols (SWP\_S)} & Maps operators to others (e.g., `$+$` $\rightarrow$ `$\times$`). & \texttt{123+10} $\rightarrow$ $123 \times 10 = 1230$. \\
\midrule
Numeric & \texttt{swap\_digits (SWP\_D)} & Swaps two digits (e.g., `1`$\leftrightarrow$`2`) in query and result. & If `1`$\leftrightarrow$`2`, \texttt{123+10} $\rightarrow$ \texttt{213+20}=233. Result 233 $\rightarrow$ 133. \\
& \texttt{reverse\_place\_values (REV\_P)} & Reverses digits in query numbers and the final result. & \texttt{123+10} $\rightarrow$ \texttt{321+1}=322. Result 322 $\rightarrow$ 223. \\
& \texttt{rotate\_digits (ROT\_D)} & Cyclically shifts digits in query numbers and the result. & \texttt{123+10} $\rightarrow$ \texttt{312+1}=313. Result 313 $\rightarrow$ 331. \\
\midrule
Operations & \texttt{add\_constant (ADD\_C)} & Adds a constant to the expression's total value. & With const=+5, \texttt{123+10} $\to 133+5=138$. \\
& \texttt{negate\_result (NEG\_R)} & Negates the expression's final result. & \texttt{123+10} $\to -(133) = -133$. \\
& \texttt{modular\_arithmetic (MOD\_A)} & Applies a modulo to the expression's result. & With mod 11, \texttt{123+10} $\to 133 \pmod{11} = 1$. \\
\midrule
Other/Advanced & \texttt{trig\_substitution (TRIG\_S)} & Substitutes the first number with a trig value (e.g., $\cos(60^\circ)$). & \texttt{123+10} $\to \cos(60^\circ)+10 = 10.5$. \\
& \texttt{change\_base (CHG\_B)} & Converts the final result to a new number base (e.g., base-8). & Result 133 $\rightarrow$ $205_8$. \\
\bottomrule
\end{tabular}%
}
\vspace{-1.5ex}
\caption{Overview of the mathematical transformations, each defining a counterfactual rule to test reasoning.}
\label{tab:transformations}
\vspace{-3ex}
\end{table*}

\section{Related Work}
A variety of benchmarks have been developed to assess mathematical reasoning abilities: MATH and MATH500 \citep{hendrycksmath2021,lightman2023let-math500} for competition problems; GSM8K \citep{cobbe2021gsm8k} for multi-step high school math; and ASDiv \citep{miao-etal-2020-diverse} for lexically and structurally diverse arithmetic word problems. 

However, it has become a concern that performance on these test sets comes from memorization. SVAMP \citep{patel-etal-2021-nlp} applied question variations and found that solvers rely on shallow heuristics. 
GSM1K \citep{zhang2024careful} mirrored the style and difficulty of GSM8K while ensuring question novelty to reveal a score drop. 
GSM-Symbolic \citep{mirzadeh2024gsm} argued that LLMs' mathematical reasoning may be attributed to matching training data, or even worse, memorizing benchmarks. \citet{huang2025math} used hard perturbations of MATH to show memorization as a major bottleneck in LLMs. 
An effective mechanism to counter memorization is using absolutely unseen tests, e.g.\ problems from yearly contests like AIME and IMO \citep{balunovic2025matharena}. However, constructing such data requires continuous manual updates. Our method automatically constructs test questions and answers from pre-defined counterfactual rules at test time. We use random seeds to set transformation parameters and select in-context examples, creating a dynamic evaluation resistant to contamination.

GSM-Symbolic dynamically evaluates arithmetic math using test-time instantiation; our work can be seen as building an inductive layer on top to test reasoning rather than purely mathematical expressions. Other dynamic benchmarks include: AntiLeakBench \citep{wu-etal-2025-antileakbench}, which constructs questions based on real-world knowledge updated after a model's training cutoff, 
and MATHHAY \citep{wang2024mathhayautomatedbenchmarklongcontext}, which generates math reasoning tasks from real-world documents. 
These approaches require advanced tools like LLMs in test creation and focus on the application of existing knowledge, whereas our work is built procedurally and simply using counterfactual rules. 

\citet{cheng2024inductivedeductiverethinkingfundamental} found that on counterfactual tasks, LLMs are surprisingly good at inductive reasoning (inferring rules) but struggle with deductive reasoning (applying rules). Similarly, \citet{qiu2024phenomenal} showed that while LLMs are effective hypothesis generators, they struggle to follow their own induced rules. Both studies highlight the gap between rule induction and application. Our work extends this with an automatically generated dataset that uses in-context examples to create novel, counterfactual math rules. By including optional hints, our benchmark enables a fine-grained analysis of inductive reasoning and whether models can favor new rules over memorized knowledge.

\citet{wu-etal-2024-reasoning} evaluate LLMs on counterfactual variants of tasks, including arithmetic, finding large drops compared to default versions. They state the counterfactual rule in the prompt, mainly testing rule execution (deduction). We additionally hide the rule, pair induction with deduction on identical problems, scale in-context examples up to 512 shots, and generate test sets procedurally from a seed. \citet{sun-etal-2024-itd} improve induction by fine-tuning on deductively generated data, evaluating on held-out examples of task types seen in training. Our fine-tuning experiments instead test transfer to held-out transformations, where supervised fine-tuning overfits to the seen rules (Section~\ref{sec:finetune}).

\section{The \mathemagic{} Framework}
\subsection{Benchmark Design and Creation}
To assess mathematical reasoning while avoiding (and revealing) memorization, we introduce \mathemagic{}, a framework that dynamically and procedurally generates counterfactual arithmetic tasks at test time, in which the rules of arithmetic are systematically altered. Consequently, a model 
cannot rely on memorized math operations and symbols and must instead reason from examples. Our framework is created in three stages: \textit{question extraction}, \textit{counterfactual transformation}, and \textit{test generation}. 

First, we extract ground truth mathematical expressions from GSM8K \citep{cobbe2021gsm8k}. Specifically, we parse the content within \texttt{<<...>>}, which represents calculation steps (e.g. $15 + 4 = 19$). By using these equations, our benchmark is based on realistic arithmetic problems. Most are single-operation expressions, though some contain two or more operations (e.g.\ $8-5-3$). Since these equations are sourced from a popular dataset, it increases the likelihood that a model has memorized the result of these common calculations. This is \textit{intentional} in our design, allowing us to analyze memorization versus reasoning abilities. When we apply counterfactual transformations, recalled answers will likely be incorrect. To succeed, the model must go beyond simple recall and reason about the transformation or use the explicit rules provided in the prompt.

Next, we define nine counterfactual transformations, which are used to modify these expressions. As shown in Table~\ref{tab:transformations}, these can be categorized into four groups based on nature: 1) \textbf{fundamental} rules alter core arithmetic symbols (e.g., swapping $+$ with $\times$); 2) \textbf{numeric} representation of numbers themselves are transformed (e.g., reversing digits); 3) \textbf{operational} functions apply a global change to the expression (e.g., adding a constant); 4) \textbf{other/advanced} concepts introduce more complexity, such as changing the number base or using trigonometric substitution. 

Finally, test questions are generated at test time using the transformations. This is controlled by a random seed for both variability and reproducibility. For each test instance, an expression from the extracted GSM8K set is randomly selected according to a chosen seed; the specific parameters of that transformation, such as the constant added or base selected, are also determined by the seed. When applicable, few-shot examples with the same transformation and parameters are randomly generated.

\subsection{Inductive and Deductive Reasoning}
Each test instance can be presented in two distinct settings to probe different reasoning skills. The \textbf{inductive} setting provides only examples, requiring the model to infer the transformation rule on its own. The \textbf{deductive} setting provides an explicit hint describing the rule alongside the examples. Note that in $k>0$-shot scenarios, the deductive setting acts as a hybrid of induction and deduction, providing both the hint and examples. Examples of these prompt configurations are in Appendix~\ref{sec:prompts}.

\begin{table*}[t]
\centering
\footnotesize
\resizebox{0.725\textwidth}{!}{%
\begin{tabular}{lrrrrrrrrrrrr}
\toprule
 \multirow[b]{2}{*}{\diagbox[width=22ex,height=1.75em]{Model}{Shots}} & \multicolumn{4}{c}{Avg. Acc. (\%)} & \multicolumn{4}{c}{SD (across seeds)} & \multicolumn{4}{c}{SD (across prompts)} \\
 \cmidrule(lr){2-5}
 \cmidrule(lr){6-9}
 \cmidrule(lr){10-13}
 & 0 & 8 & 64 & 512 & 0 & 8 & 64 & 512 & 0 & 8 & 64 & 512 \\
\midrule
Qwen2.5 0.5B & 3.9 & 5.0 & 8.0 & 1.8 & 1.3 & 0.9 & 0.9 & 0.2 & 2.6 & 5.8 & 1.9 & 1.2 \\
Qwen2.5 1.5B & 17.4 & 19.4 & 19.4 & 16.9 & 1.0 & 0.9 & 1.3 & 1.7 & 1.0 & 0.4 & 0.4 & 2.9 \\
Qwen2.5 3B & 19.0 & 29.4 & 31.7 & 31.0 & 1.3 & 1.7 & 2.0 & 0.8 & 1.9 & 3.6 & 3.8 & 4.4 \\
Qwen2.5 7B & 19.0 & 29.1 & 34.7 & 38.7 & 1.6 & 2.9 & 1.8 & 2.9 & 0.1 & 3.0 & 2.5 & 3.0 \\
Qwen2.5 14B & 19.7 & 33.0 & 45.2 & 49.4 & 1.7 & 1.3 & 1.3 & 2.0 & 0.1 & 1.8 & 4.2 & 4.4 \\
Qwen2.5 32B & 19.7 & 42.3 & 51.2 & 57.7 & 1.7 & 2.0 & 1.3 & 2.7 & 0.0 & 3.5 & 1.1 & 1.1 \\
Qwen2.5 72B & {19.8} & 40.9 & 46.9 & 51.3 & 1.7 & 2.4 & 1.4 & 0.8 & 0.0 & 1.9 & 3.3 & 4.0 \\
Llama3 8B & 18.4 & 28.5 & 32.7 & - & 1.5 & 2.0 & 1.2 & - & 9.0 & 13.5 & 14.3 & - \\
GPT-4o-mini & {19.8} & 40.3 & 44.3 & 47.3 & 1.7 & 2.1 & 1.8 & 2.0 & 0.0 & 0.2 & 1.7 & 4.4 \\
GPT-5-nano & 18.9 & 61.5 & 64.3 & 65.3 & 1.1 & 1.8 & 2.1 & 2.0 & 0.4 & 0.5 & 0.9 & 1.9 \\
GPT-5 & {19.8} & {73.7} & {83.8} & {87.7} & 1.7 & 1.8 & 1.2 & 1.1 & 0.0 & 0.4 & 0.9 & 1.4 \\
\bottomrule
\end{tabular}
}%
\vspace{-1.5ex}
\caption{Inductive performance (Avg. Acc.) and standard deviation (SD) across seeds or prompt variants, under 0, 8, 64, and 512 shots. Model performance is stable even when the test sets are dynamically constructed.}
\label{tab:stability}
\vspace{-3ex}
\end{table*}

\subsection{Feature Highlights and Extensibility}
Our benchmark design offers several advantages, including reproducibility, robustness to overfitting, extensibility, and comprehensiveness, which tackle a few pain points of existing math test sets. These features ensure the benchmark's long-term relevance as a stable tool for measuring generalization. 

The procedural and seeded generation ensures both reproducibility and robustness to memorization. By using the same random seed, a test can be fully replicated; using different random seeds creates a moving target (test set), making the benchmark resistant to contamination and memorization. 

Our modularized design makes the benchmark easily customizable and extensible. In principle, data can be extracted and seeded from other mathematical sources, although we only evaluate arithmetic expressions. It is also easy to implement new transformations for the test, which can be utilized to adjust the difficulty or further avoid overfitting to the existing transformations. Additionally, our benchmark can be used to test both inductive and deductive reasoning, enabling a more fine-grained analysis of the model's reasoning abilities. 

Our use of basic arithmetic serves as a primary proof of concept. Because these operations are common in training data, they are the most likely to be memorized. Demonstrating that models struggle in such simple cases provides stronger evidence of reasoning failure than more complex tasks might. Our findings therefore concern arithmetic; whether they hold for algebra, multi-step word problems or proofs is left to future work. More broadly, this procedurally generated methodology is extensible to any domain defined by a formal grammar, such as linguistic or logical reasoning.

\section{Experiments and Findings}
Our experiments serve to validate \mathemagic{} and to study the reasoning ability of LLMs under arithmetic transformations. We assess open-source LLMs: Llama3 8B Instruct \citep{grattafiori2024Llama3herdmodels}; the Qwen2.5 Instruct series \citep{qwen2025qwen25technicalreport} from 0.5B to 72B, and proprietary models: GPT-4o-mini \citep{openai2024gpt4omini}, GPT-5-nano, GPT-5 \citep{openai2025gpt5} in ``thinking'' mode. 

We generated five versions of test sets using five different random seeds, each containing 405 unique problems covering our nine mathematical transformations. In all experiments, we use greedy decoding. The model output is extracted from the LaTeX \verb|\boxed{}|
and considered correct if within a relative and absolute tolerance of $10^{-4}$ of the reference value. Unless otherwise specified, all experiments use a prompt asking for step-by-step reasoning before answering (Appendix~\ref{sec:prompts}). We run 0- to 512-shot prompting, but Llama3 8B is limited to a maximum of 256 shots due to context size.

First of all, to validate our benchmark's reliability, we assessed the stability of models' inductive performance against two variations: five different random seeds to measure robustness to numerical changes; and four distinct instruction prompts (Appendix~\ref{sec:prompts}) to measure sensitivity to phrasing. LLMs demonstrated consistent performance across both seed and prompt variations, as shown in Table~\ref{tab:stability}, with Llama3 8B as an exception, exhibiting instability in response to different prompts, which suggests its reasoning is more dependent on input wording. Overall, the observed stability justifies our use of a dynamically constructed benchmark.

\subsection{Does induction performance improve with more in-context examples?}
A foundational capability of modern LLMs is in-context learning, where a model adapts to a new task based on examples provided directly in its prompt. First, we establish a baseline by investigating how model performance on our counterfactual tasks scales with the number of examples. We evaluate LLMs inductively, where they must infer the transformation rule from examples alone. We tested each model from 0 to 512 examples. To provide a robust measure of overall capability, the reported accuracy is averaged across all transformation types and the five random seeds.

\begin{figure}[t]
\centering\footnotesize
  \includegraphics[width=\columnwidth]{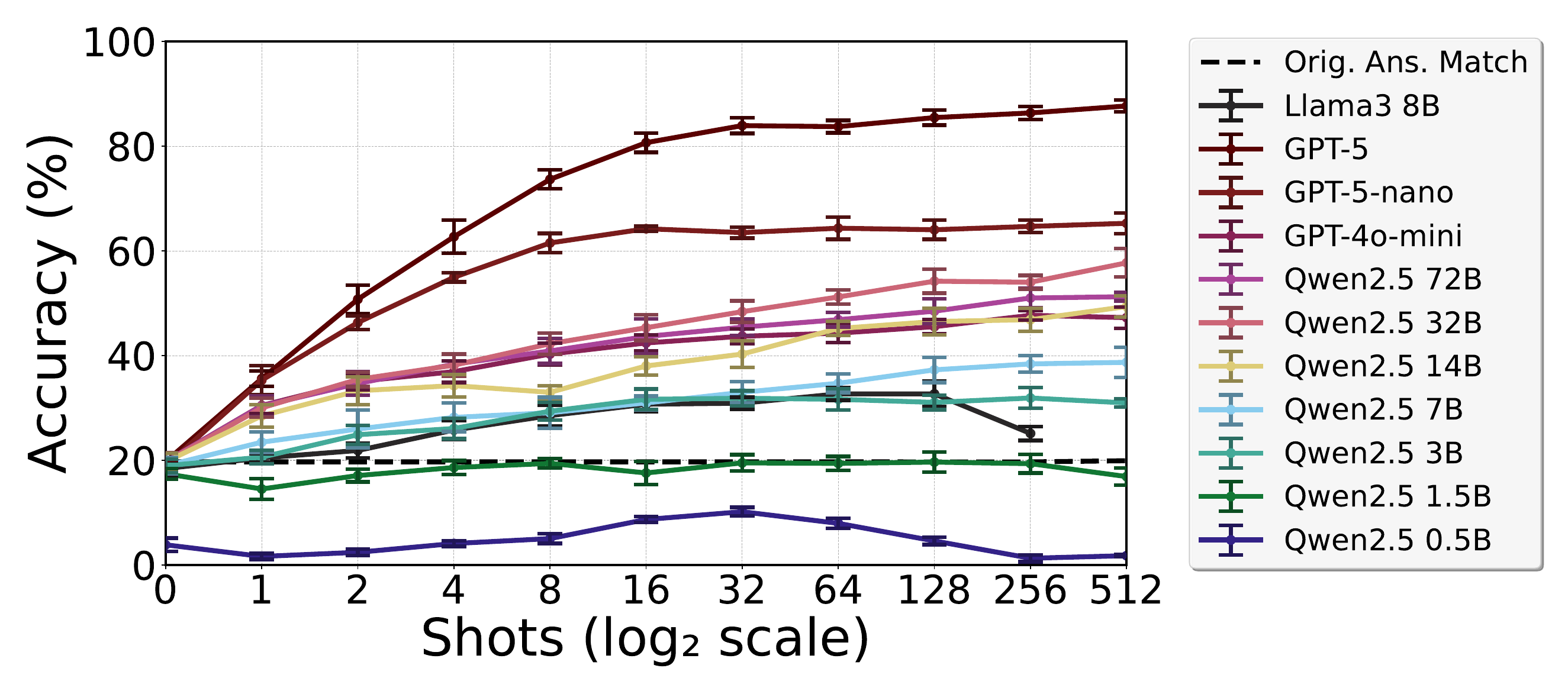}
  \vspace{-5.5ex}
  \caption{Performance of in-context learning (inductive reasoning) across shots.
  }
  \label{fig:context_impact}
  \vspace{-1.5ex}
\end{figure}

\begin{figure}[t]
\centering\small
  \includegraphics[width=\columnwidth]{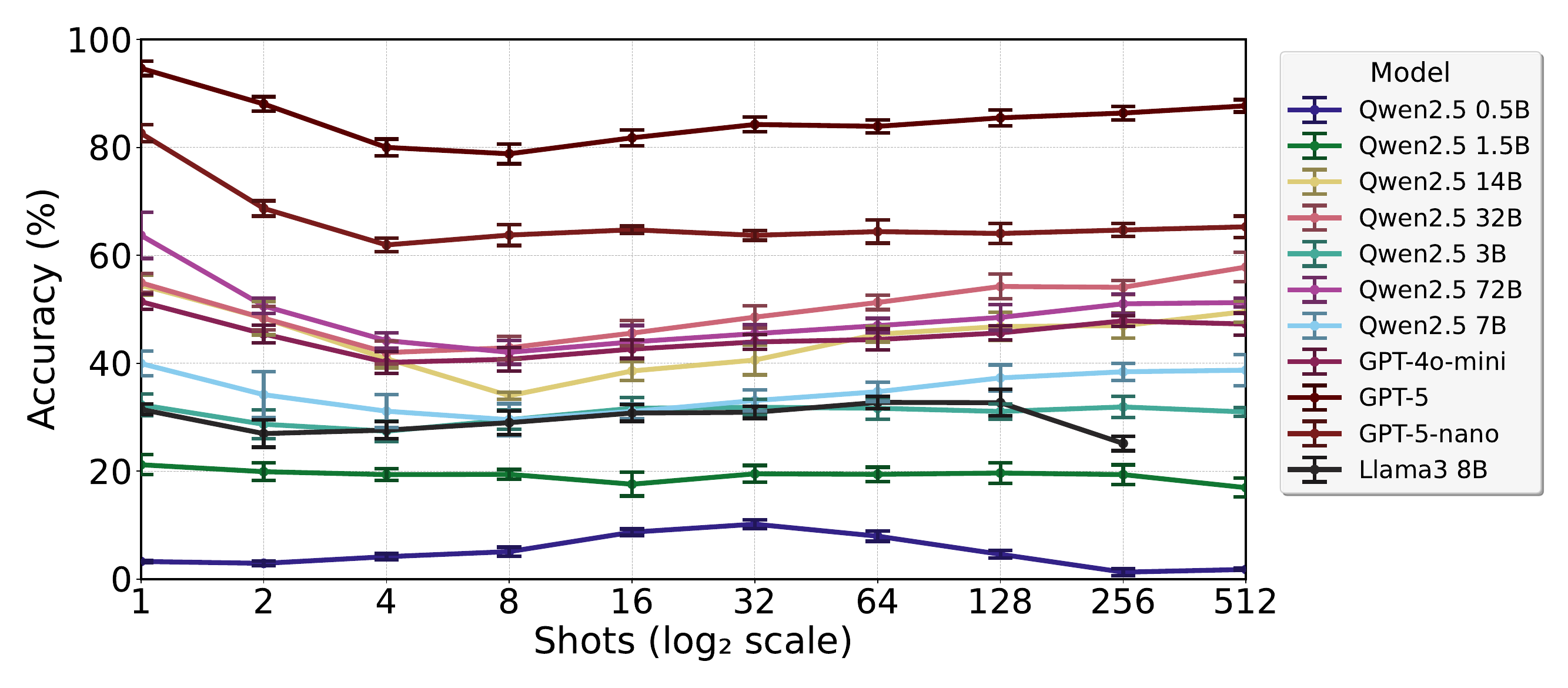}
  \vspace{-5ex}
  \caption{In-context learning performance with verifier applied. Correct answers remain correct; incorrect answers are reclassified as correct only if consistent with the given examples and question.
  }
  \label{fig:verifier}
  \vspace{-2ex}
\end{figure}

Figure~\ref{fig:context_impact} shows that scores improve with more examples. Zero-shot results start at around 20\%, which is the proportion of questions where the standard-arithmetic answer is coincidentally correct (Orig. Ans. Match), e.g.\ swapping $+$ and $\times$ leaves $2+2=4$ unchanged. After that, performance varies with model scale and the number of examples. Reasoning models, GPT-5 and GPT-5-nano, show a clear positive trend, rising steadily with more examples before plateauing around the 32 shots. We hypothesize this is due to the models' use of ``thinking tokens'' to process the novel rules, although it merits a more in-depth analysis. Conversely, the smaller models (Qwen2.5 1.5B and 0.5B) failed to learn from the context. We also identify a failure mode in Llama3 8B, which improved moderately before dropping sharply after 128 shots, indicating that excessive examples can be detrimental.

\paragraph{Unchanged answers.} We keep the coincidentally correct questions for three reasons. First, the transformation is still applied correctly, the answer just happens to match. Second, the rate differs a lot between transformations (0\% for NEG\_R, 65\% for ROT\_D), so removing them would change the distribution of transformation types. Third, removing them would bias models to assume the answer is never the standard one.. A single question cannot tell rule-following apart from standard arithmetic, but its examples can: at 512 shots, 93.8\% of these questions come with an example that contradicts standard arithmetic. 

In any case, removing these questions makes the task harder (Appendix~\ref{app:variants}, Figures~\ref{fig:variant1}--\ref{fig:variant4}, panel~b). In induction, zero-shot accuracy drops to at most 0.4\% for every model, and Qwen2.5 72B at 512 shots falls from 51.3\% to 44.1\%. Deduction is barely affected (Qwen2.5 72B at zero shots scores 81.7\% both before and after), and the gap between 512-shot induction and zero-shot deduction widens from 30.4 to 37.6 points. The unfiltered numbers therefore understate how hard the task is.

\begin{table}[t]
\centering\footnotesize
\setlength{\tabcolsep}{3pt}
\resizebox{0.65\linewidth}{!}{%
\begin{tabular}{lrrr}
\toprule
\multirow{2}{*}{\diagbox[width=22ex,height=1.75em]{Model}{Shots}} & \multicolumn{3}{c}{Avg. Acc. (\%)} \\
\cmidrule(lr){2-4}
& 0  & 8  & 64  \\
\midrule
Qwen2.5 1.5B Instruct & 17.4 & 19.4 & 19.4 \\
Qwen2.5 1.5B {Math} Instruct & 19.7 & 19.8 & 19.7 \\
\midrule
Qwen2.5 7B Instruct & 19.0 & \textbf{29.1} & \textbf{34.7} \\
Qwen2.5 7B {Math} Instruct & 19.6 & 25.3 & 20.4 \\
\bottomrule
\end{tabular}
}%
\vspace{-1.5ex}
\caption{Performance of Qwen2.5 Instruct and Qwen2.5 Math Instruct on inductive reasoning in \mathemagic{}.}
\label{tab:math-instruct-vs-instruct}
\vspace{-2ex}
\end{table}

\subsection{What about math LLMs?}
We compare general-purpose and math-adapted models: Qwen2.5 Instruct and Qwen2.5 Math Instruct \citep{yang2024qwen2}. Table~\ref{tab:math-instruct-vs-instruct} shows that both Qwen2.5 1.5B models stay at the 20\% mark, while the 7B general-purpose model improves with more shots and the Math version does not. Math-specialised models are therefore less flexible at following unfamiliar arithmetic rules on our benchmark, tending to fall back on standard arithmetic. Our results raise the question of whether adapting to math is a genuine strategy for enhancing (mathematical) reasoning.

\subsection{Are there other valid explanations for the same answer?} 
\label{sec:verifier}
A limitation of scoring inductive responses strictly against a single reference rule is that, especially with fewer examples, there can exist multiple rules for the same examples. A model's answer may be judged wrong because it followed a different but equally valid explanation. To check how much this affects our results, we build a verifier: for responses whose boxed answer disagrees with the reference, we take the model's own explanation (without the question or examples), prompt Gemini 3.5 Flash \citep{geminiflash35} to turn it into a Python function, and run this function over all the in-context examples shown. If it reproduces every example and the model's own answer to the question, we count the response as correct.

A function only earns credit if it fits every example and gives the model's answer, so a credited function is a valid explanation of that answer. To check that these functions reflect the models' own reasoning, we run two checks. First, for GPT-5 and Qwen2.5 72B at 1 and 512 shots, Gemini's function reproduces the model's own answer in 92--99\% of cases. Second, we manually audited 100 translations sampled across all models and shot counts: 96\% faithfully implement the model's stated reasoning. Often, however, the model's reasoning gave no clear rule (e.g.\ guessing an answer by analogy to a nearby example); we count a translation of such reasoning as faithful if it reproduces the model's answer. Since most translation errors can only cause missed credit, verified accuracy should be read as an approximate lower bound on how often other valid explanations exist, rather than an exact count.

Figure~\ref{fig:verifier} shows the accuracy, where incorrect answers that have been verified are added to already correct ones. Unsurprisingly, we get increased accuracy as there are multiple valid explanations. The size of this gap is not constant across shots: it is largest at low shot counts (up to 59 points for GPT-5 at one shot) and shrinks quickly, to almost nothing from 16 shots onwards. With only a few examples, many different rules remain consistent with what has been shown, so a model's own explanation is likely to be judged correct even when it does not match the intended rule. As more examples are added, this ambiguity narrows with fewer alternative explanations, with most of this narrowing occurring by around 8 shots. Beyond this, further examples narrow the space of valid rules only a little more, since most of the narrowing has already happened; what they add instead is more evidence for whichever rule is consistent with all of them, so gains in accuracy come from finding the intended rule rather than from alternative explanations.

\subsection{Is rule induction or deduction harder?}
\begin{figure}[t]
\centering\small
  \includegraphics[width=0.95\columnwidth]{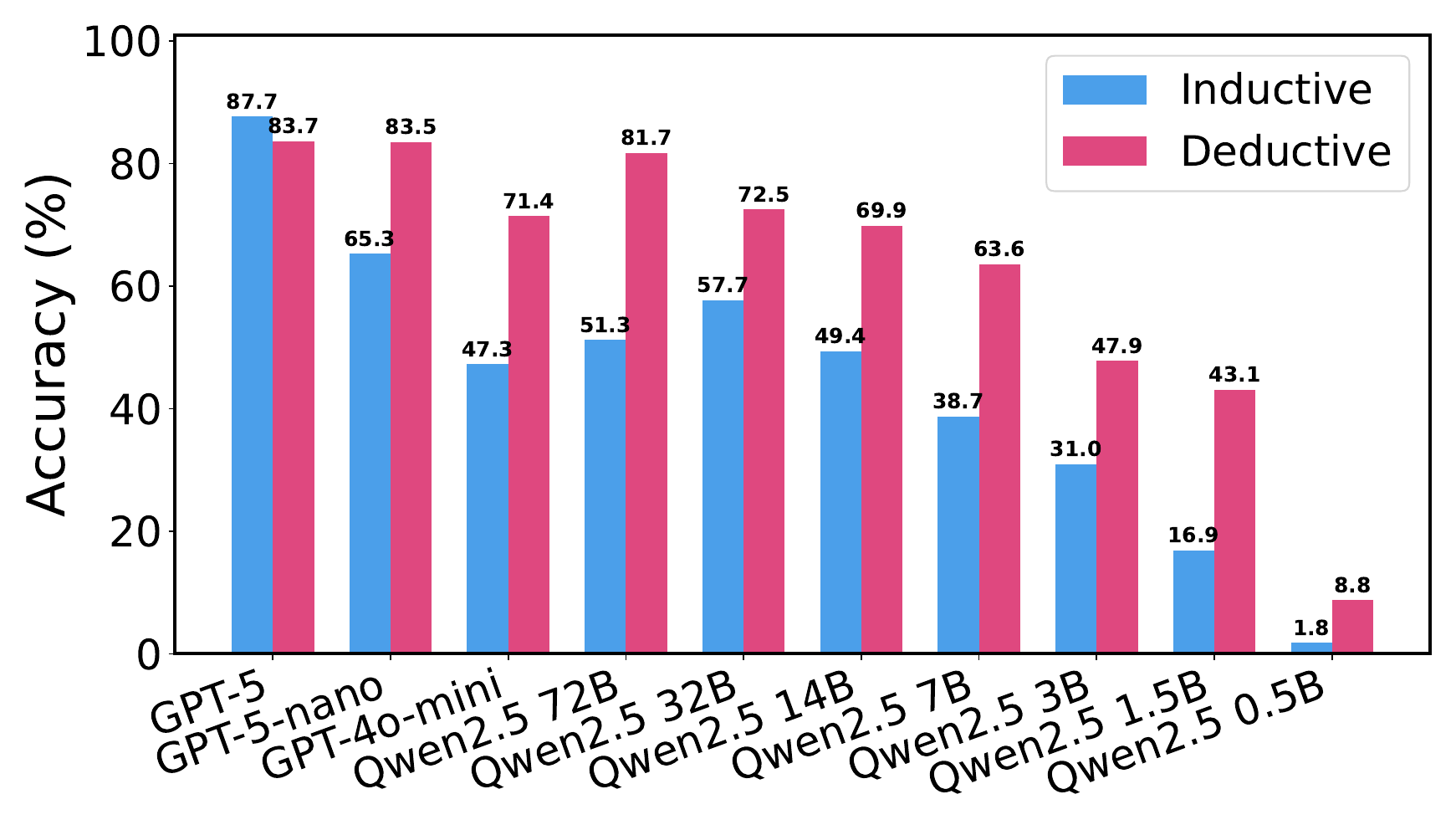}
\vspace{-3ex}
  \caption{Model performance with inductive at 512 shots, and deductive at 0 shots.
  }
  \label{fig:inductive_deductive}
\vspace{-1.5ex}
\end{figure}

Each instance can be posed in one of two settings. In the inductive setting, the model receives only in-context examples and must infer the transformation itself; in the deductive setting, an explicit natural-language rule is also supplied. Solving a task inductively therefore requires two distinct steps: correctly inferring the rule, and correctly executing it. A wrong final answer alone does not reveal which of these has failed. We compare induction at 512 shots, and deduction at 0 to give the inductive setting the highest number of examples, while deduction just has to rely on the rule.

\begin{table}[t]
\centering
\footnotesize
\setlength{\tabcolsep}{3pt}
\resizebox{\linewidth}{!}{%
\begin{tabular}{lrrrrrrrr}
\toprule
\multirow[b]{2}{*}{\diagbox[width=22ex,height=1.75em]{Model}{Shots}}
& \multicolumn{4}{c}{Induction (\%)} & \multicolumn{4}{c}{Deduction (\%)} \\
\cmidrule(lr){2-5}\cmidrule(lr){6-9}
& 1 & 8 & 64 & 512 & 1 & 8 & 64 & 512 \\
\midrule
Llama3 8B & 15.0 & 1.2 & 1.1 & -- & 86.6 & 83.9 & 81.3 & -- \\
GPT-5 & 92.6 & 19.7 & 1.2 & 0.0 & 97.9 & 98.3 & 99.1 & 99.2 \\
GPT-5-nano & 77.3 & 6.9 & 0.3 & 0.7 & 84.5 & 85.9 & 75.2 & 83.6 \\
GPT-4o-mini & 31.3 & 3.5 & 4.3 & 2.7 & 92.4 & 79.4 & 75.0 & 80.3 \\
Qwen2.5 72B & 49.3 & 6.6 & 3.1 & 0.6 & 97.3 & 85.0 & 84.5 & 91.2 \\
Qwen2.5 32B & 35.9 & 4.1 & 4.5 & 1.5 & 95.2 & 79.6 & 71.9 & 78.5 \\
Qwen2.5 14B & 38.9 & 24.8 & 12.5 & 4.0 & 92.7 & 56.3 & 65.6 & 81.8 \\
Qwen2.5 7B & 22.9 & 5.2 & 2.1 & 0.9 & 94.1 & 84.1 & 84.4 & 88.6 \\
Qwen2.5 3B & 16.0 & 2.2 & 0.4 & 0.5 & 91.9 & 89.4 & 94.0 & 94.1 \\
Qwen2.5 1.5B & 10.6 & 1.2 & 0.1 & 0.6 & 65.3 & 72.6 & 90.0 & 76.9 \\
Qwen2.5 0.5B & 13.4 & 1.1 & 0.5 & 2.4 & 12.2 & 31.9 & 46.6 & 8.7 \\
\bottomrule
\end{tabular}
}%
\vspace{-1.5ex}
\caption{Inductive versus deductive performance on answers that were incorrect on the original transformation.}
\label{tab:induction-deduction-by-shots}
\vspace{-2ex}
\end{table}

Figure~\ref{fig:inductive_deductive} shows a coarse comparison between the two settings at a fixed 512 shots, averaged across all nine transformations. Most models with the rule and zero examples can beat a model with 512 examples but no rule. For instance, GPT-4o-mini's zero-shot deductive accuracy of 71.4\% is considerably higher than its 47.3\% accuracy on the 512-shot inductive task. On the other hand, GPT-5 learned to use many-shot examples, getting a 512-shot inductive accuracy of 87.7\% compared to its zero-shot deductive accuracy of 83.7\%.

To separate the components of inductive failure, we apply the verifier from Section \ref{sec:verifier}. For each such response, we apply the transformation from the models explanation in two ways: whether executing this self-proposed rule reproduces the given examples (deduction, i.e.\ can the model correctly apply a rule it believes in), and whether this rule matches the ground-truth transformation (induction, i.e.\ did it identify the correct rule at all). Because this analysis only considers responses that were already incorrect, induction accuracy is naturally low overall. Table~\ref{tab:induction-deduction-by-shots} shows that deduction accuracy remains high across all shot counts, while induction accuracy stays low and declines further as more examples are provided, since more examples leave fewer rules consistent with all of them, making a correct match harder to reach by chance. Deduction, by contrast, only asks whether the model applies its own stated reasoning correctly, so it shows no such decline. Induction is therefore more difficult than deduction.

\paragraph{Are the examples treated as noise?} A possible explanation for poor induction is that models dismiss counterfactual examples as noise and fall back to standard arithmetic. To test this, we build an incoherent variant of the task: each in-context example applies a \emph{different} transformation with its own parameters, so no single rule explains them (405 questions). If models treated our examples as noise, they would behave the same way on both versions. They do not. Qwen2.5 14B answers with standard arithmetic far more often when the examples are incoherent: 21.0\% versus 11.1\% at 8 shots, and 44.5\% versus 14.5\% at 64 shots, both on questions where standard arithmetic gives a wrong answer.\footnote{On both the coherent and incoherent sets, a question is excluded if one randomly chosen shown rule maps it to its standard-arithmetic answer.} Format errors also rise, from 16.3\% to 28.4\% at 8 shots and from 3.6\% to 10.9\% at 64. Models therefore treat a consistent counterfactual rule as meaningful, and their inductive failures come from inferring the rule rather than ignoring the examples.

\subsection{What makes certain transformations difficult for models?}
Our benchmark's diverse set of transformations, from simple operational changes (e.g., \texttt{NEG\_R}) to complex procedural manipulations (e.g., \texttt{REV\_P}), enables us to investigate if some transformations are inherently more difficult. To understand \textit{why} certain tasks are harder, we analyze the specific \textit{nature} of a model's failures. This combined analysis of task difficulty and error types provides a deeper insight into models' reasoning limitations.

To investigate whether all transformations pose equal difficulty, we analyze the performance of Qwen2.5 72B, on a per-task basis. We compare accuracy in both inductive and deductive settings with a large context of 512 shots. To understand why models fail, we also inspect the distribution of errors in the inductive and deductive setting, which we categorize into four types: \textbf{correct}; \textbf{reversion error} (defaulting to standard math); \textbf{novel error} (an incorrect attempt to apply the new rule); and \textbf{format error} (unparsable output). By comparing the error breakdown at zero, eight, and 512 shots, we can observe how failure patterns evolve.

\begin{figure}[t]
\centering\small
  \includegraphics[width=\columnwidth]{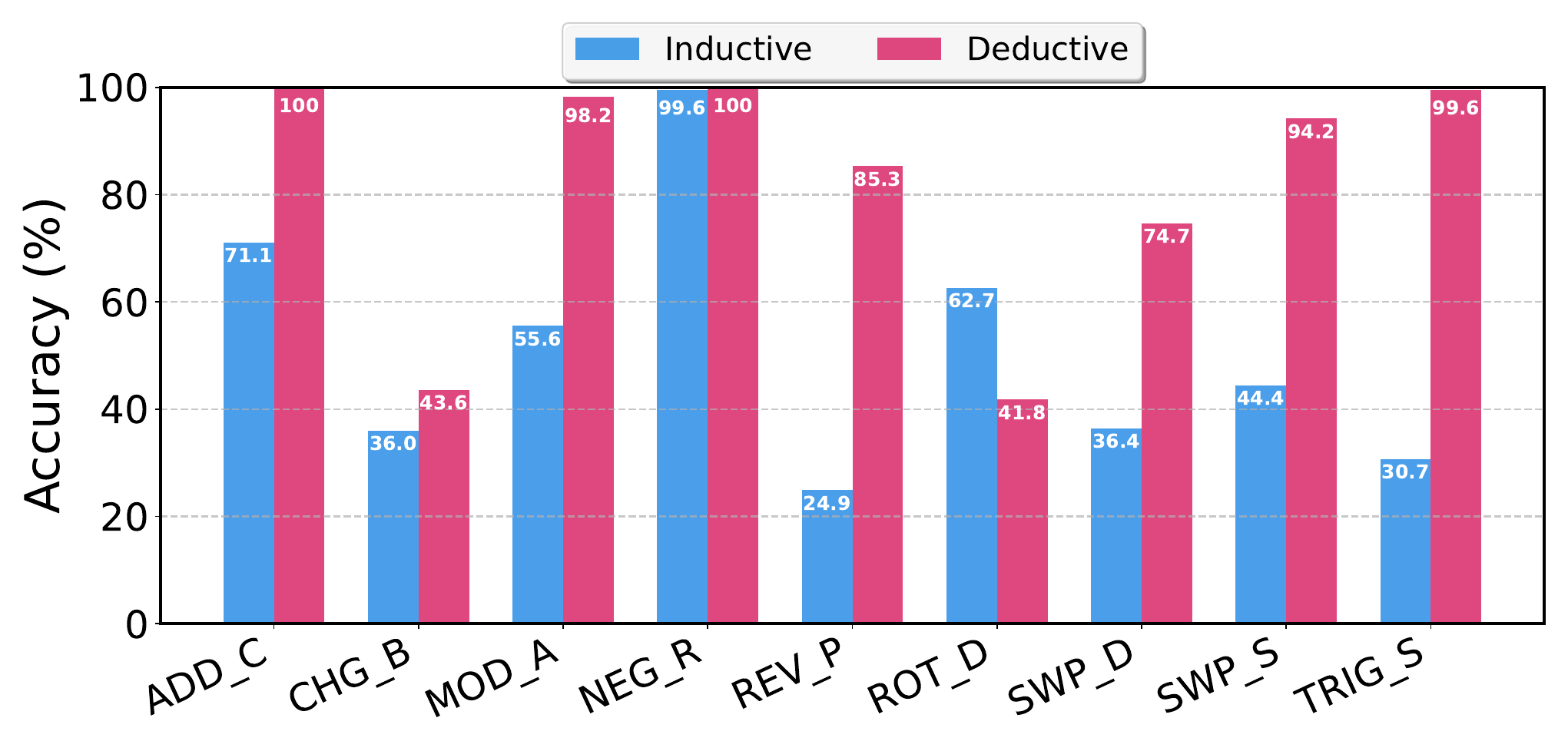}
\vspace{-4.5ex}
  \caption{Qwen2.5 72B results by transformation type, with inductive and deductive at 512 shots.
  }
\label{fig:transform_diff}
\vspace{-2ex}
\end{figure}

\label{sec:failure}
\begin{figure*}[t]
\centering\small
  \includegraphics[width=0.78\linewidth]{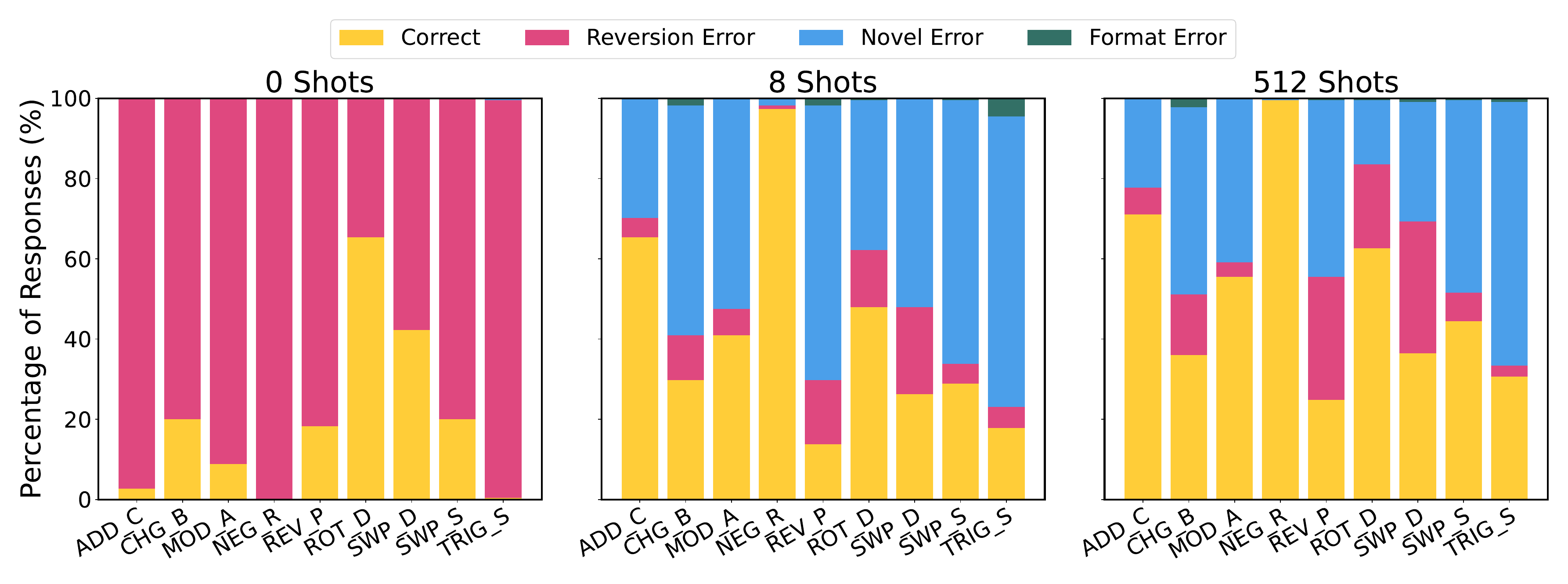}
\vspace{-3ex}
  \caption{Breakdown of response types for Qwen2.5 72B in the inductive setting at various shots. The bars show the percentage of responses that are: a correct answer, a reversion error (reverting to the original math), a novel error (incorrect under both new and original rules), or a format error.}
  \label{fig:nature_failure}
\vspace{-2ex}
\end{figure*}

An analysis of task performance (Figure~\ref{fig:transform_diff}) and error profiles (Figure~\ref{fig:nature_failure}) reveals what makes certain transformations difficult. Qwen2.5 72B excels at simple, single-step operations like \texttt{ADD\_C} and \texttt{NEG\_R} but struggles with complex procedural tasks like \texttt{CHG\_B}. Pairing induction with deduction on the same questions separates finding a rule from applying it. Qwen2.5 72B applies most rules well once given: \texttt{TRIG\_S} reaches 99.6\% with deduction versus 30.7\% with induction, and \texttt{REV\_P} 85.3\% versus 24.9\%, so its failures there come from inferring the rule. \texttt{CHG\_B} is the exception, reaching only 43.6\% even when the base is stated (36.0\% with induction). This matches \citet{wu-etal-2024-reasoning}, who find that models struggle with arithmetic in unfamiliar bases even when the base is given, so base conversion is hard to carry out as well as to discover.

The error patterns (Figure~\ref{fig:nature_failure}) show how behaviour changes with more examples. For most tasks, as more examples are provided, the rate of correct answers increases while reversion errors (defaulting to standard math) decrease. This shows the model is learning to override its pre-trained knowledge. However, this also leads to more frequent novel errors, which are incorrect attempts to apply the new rule, indicating a behavioral shift from simply ignoring the rule to attempting to apply it.

However, this learning is not uniform. Difficult tasks like \texttt{SWP\_D} and \texttt{REV\_P} show high rates of reversion even at 512 shots. This demonstrates a struggle to overcome learned patterns. The \texttt{ROT\_D} task presents a case that highlights these learning failures. In the inductive setting, its high initial accuracy is a coincidental byproduct of the task's structure, as many rotations are null operations (e.g., rotating ‘52’ by two places results in ‘52’) that make the standard answer correct by chance. 
\texttt{ROT\_D}'s performance remains static across all shot counts, which confirms that the model is not learning the procedural rule but is instead relying on the simple heuristic of giving the standard, unchanged answer. This leads to an outcome where providing an explicit rule in the deductive setting actually results in worse performance. This may be because the deductive instruction creates a bias. When told a new math rule is in play, the model may be less likely to output an answer that is the same as standard math. This expectation that a transformation must always produce a different result may lead it to force an incorrect change, causing it to perform worse than when it relied on its simple heuristic.

\begin{table}[t]
\centering
\footnotesize
\setlength{\tabcolsep}{3pt}
\resizebox{\linewidth}{!}{%
\begin{tabular}{lrrrr}
\toprule
\multirow[b]{2}{*}{\diagbox[width=16ex,height=1.5em]{Model}{Shots}}
& \multicolumn{4}{c}{Avg. Acc. (\%)} \\
\cmidrule(lr){2-5}
& 0 & 8 & 64 & 512 \\
\midrule
Qwen2.5 7B & 28.4 ($+$9.3) & 57.8 ($+$41.3) & 72.4 ($+$53.3) & 77.3 ($+$57.8) \\
Qwen2.5 14B & 31.6 ($+$11.6) & 88.4 ($+$67.6) & 99.1 ($+$63.6) & 98.7 ($+$53.8) \\
Qwen2.5 32B & 26.7 ($+$6.7) & 87.1 ($+$57.8) & 98.7 ($+$54.2) & 99.6 ($+$47.1) \\
Qwen2.5 72B & 23.1 ($+$3.1) & 96.4 ($+$67.6) & 99.6 ($+$61.3) & 99.6 ($+$55.1) \\
\bottomrule
\end{tabular}
}%
\vspace{-1.5ex}
\caption{Inductive performance across shots when operators are mapped to arbitrary tokens. Difference reported w.r.t. the corresponding counterfactual \texttt{SWP\_S}.}
\label{tab:noncounter}
\vspace{-2ex}
\end{table}

\begin{figure}[t]
  \includegraphics[width=\columnwidth]{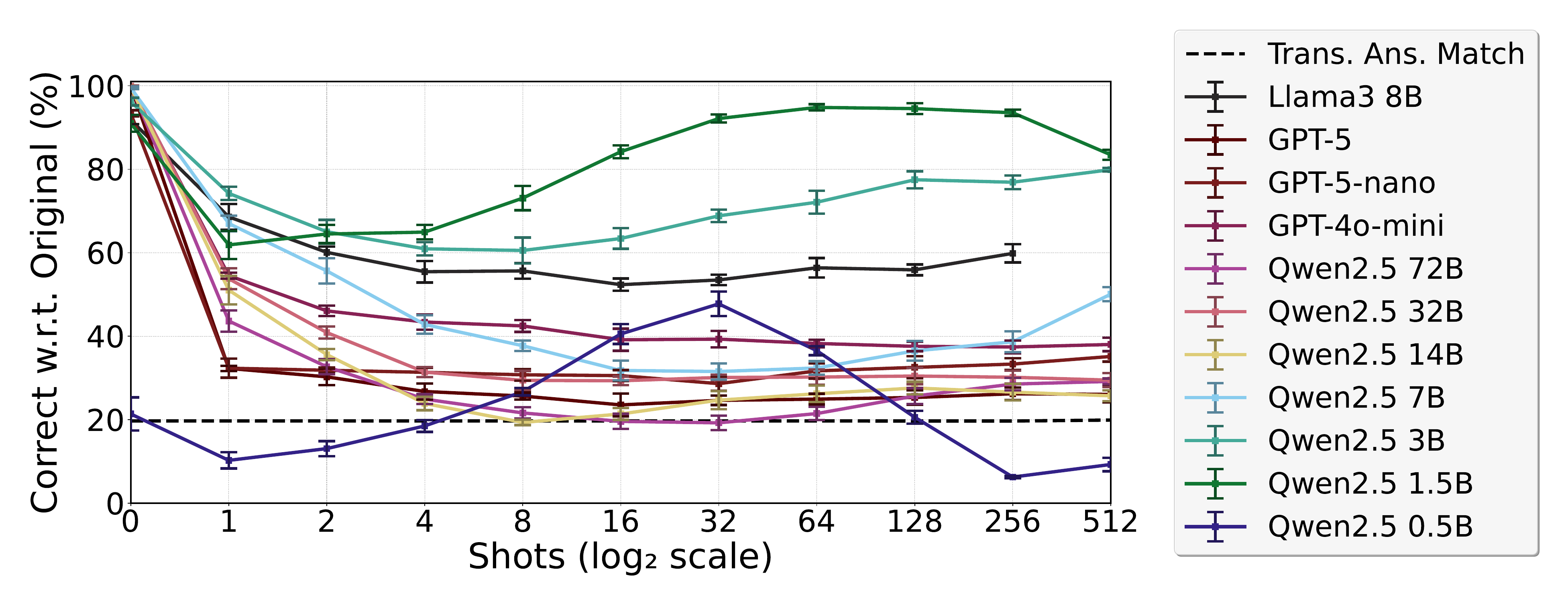}
\vspace{-4ex}
  \caption{Percentage of answers that are correct w.r.t.\ \textit{original math}, reported across shots.
  }
  \label{fig:correct_wrt_original}
\vspace{-2ex}
\end{figure}

To check whether the difficulty comes from counterfactual mathematics itself, rather than just novel mathematics, we compare \texttt{SWP\_S} against a control that maps operators to arbitrary, meaningless tokens instead of other operators (e.g, mapping 5 + 4 into 5 A 4). We mapped them to capitalised letters. Table~\ref{tab:noncounter} shows accuracy is far higher under this novel, but-non-counterfactual control, suggesting  difficulty comes  from overriding  memorized meaning, rather than learning a new mapping.

\subsection{To what extent do models recall memorized knowledge?}
While overall accuracy measures success, it cannot fully answer  whether models genuinely reason using new rules or simply fall back on their vast store of memorized standard arithmetic. This is relevant in our benchmark, where relying on standard math is a guaranteed path to failure for most problems. 

We investigate how often the answer provided by a model is correct \textit{with respect to standard arithmetic (the answer as if there is no transformation)}. A high value on this metric indicates a strong bias towards memorized knowledge, with the model ignoring the in-context examples and defaulting to its own knowledge. We plot this metric against the number of shots provided. Our analysis, shown in Figure~\ref{fig:correct_wrt_original}, indicates to what degree models struggle to suppress their ingrained knowledge when presented with counterfactual examples. 

As examples are introduced, a performance hierarchy emerges. Larger models like Qwen2.5 32B begin to adapt, but this adaptation is incomplete. They still revert to standard math at a rate higher than can be explained by chance (i.e., above the dashed \texttt{Transformed Answer Match} (Trans. Ans. Match) line). This reliance on standard arithmetic is even more pronounced 
with less capable models, which maintain a high rate of reversion to familiar patterns even with many examples. Qwen2.5 7B, for instance, shows a more volatile adaptation; its initial success at suppressing its bias is reversed as the context grows larger. 
This problem is most extreme in the smallest model, Qwen2.5 0.5B, whose erratic performance appears to be an artifact of its fluctuating ability to follow formatting instructions. Its rate of format errors drops from 61.5\% at zero shots to a low of 20.1\% at 32 shots, which is the exact point where its tendency to produce the standard math answer peaks. This suggests the model is simply confused by the context, not learning a coherent strategy. This persistent reversion shows that memorized arithmetic knowledge is a key barrier to reasoning on these tasks.

\subsection{Is inductive reasoning generalizable?} \label{sec:finetune}
Our results so far show that while models can learn from in-context examples, this ability is limited and often fails. Can the skill of inductive reasoning be taught through training? If a model is fine-tuned on our counterfactual tasks, will it learn to generalize the underlying reasoning process, or will it simply memorize the examples it has seen? To investigate this, we conducted a series of fine-tuning experiments designed to test generalization.

To test if inductive reasoning can be taught, we fine-tuned the base Qwen2.5 7B model using QLoRA \citep{dettmers2023qlora} and compare it with Qwen2.5 7B Instruct. The training data consisted of 2,500 sets of 8-shot inductive examples for each transformation, all generated using a single random seed. We conducted three experiments to test generalization: training on all transformations to test new instances of the same tasks, training on only one transformation to test skill transfer to the others, and training on eight transformations to test generalization to a new, held-out task.

\begin{figure}[t]
\centering\small
  \includegraphics[width=\columnwidth]{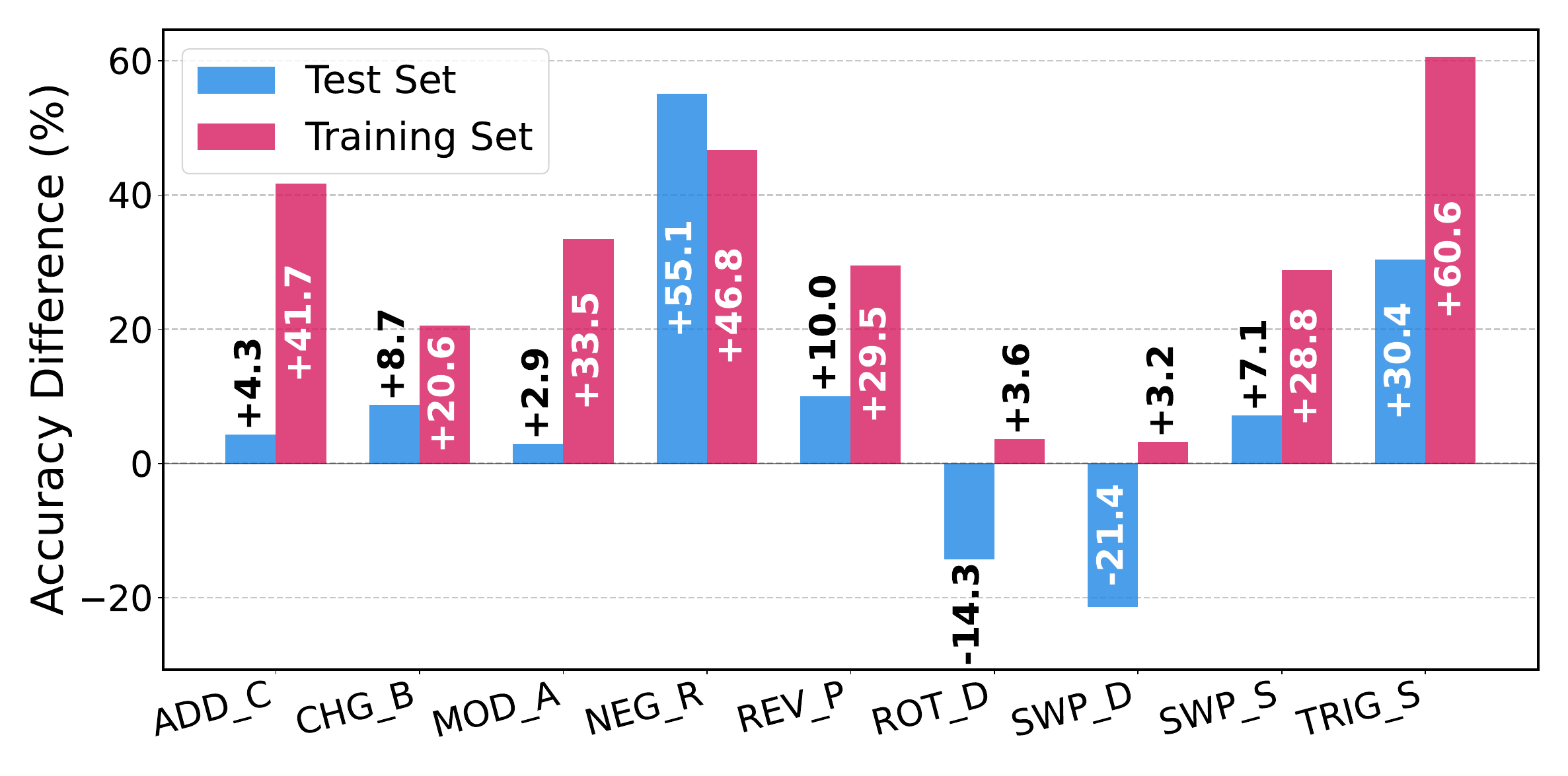}
\vspace{-4ex}
    \caption{Accuracy difference between fine-tuned Qwen2.5 7B and Qwen2.5 7B Instruct, across transformations. Fine-tuned on \textit{all} transformations; tested on training (pink) and test (blue) sets.}
  \label{fig:generalized}
\vspace{-2ex}
\end{figure}

\begin{figure}[t]
\centering\small
  \includegraphics[width=\columnwidth]{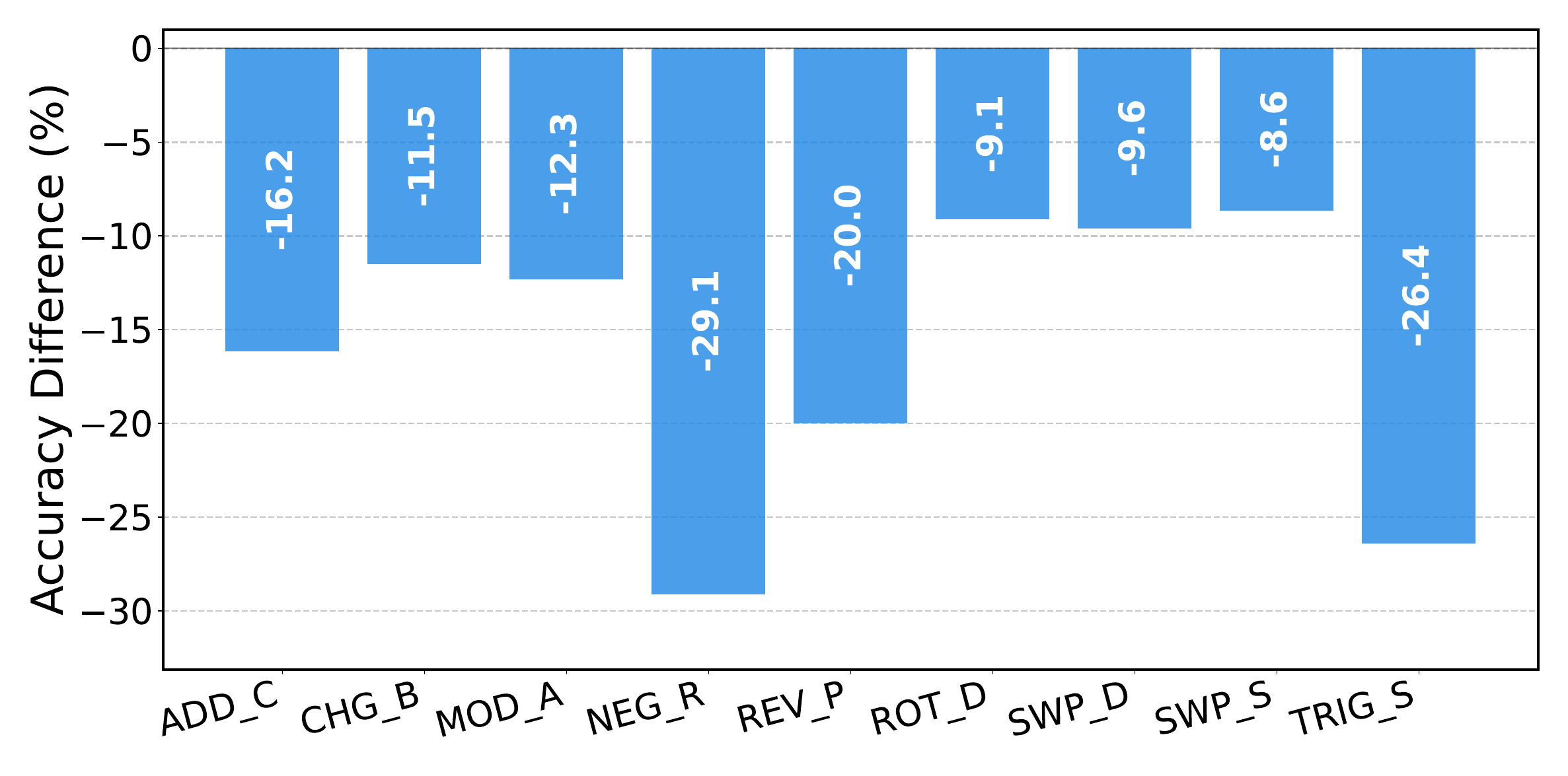}
\vspace{-4ex}
    \caption{Accuracy difference between fine-tuned Qwen2.5 7B and Qwen2.5 7B Instruct. Fine-tuned on a \textit{single} transformation and tested on the rest.}
\label{fig:train_one}
\vspace{-2ex}
\end{figure}

\begin{figure}[t]
\centering\small
  \includegraphics[width=\columnwidth]{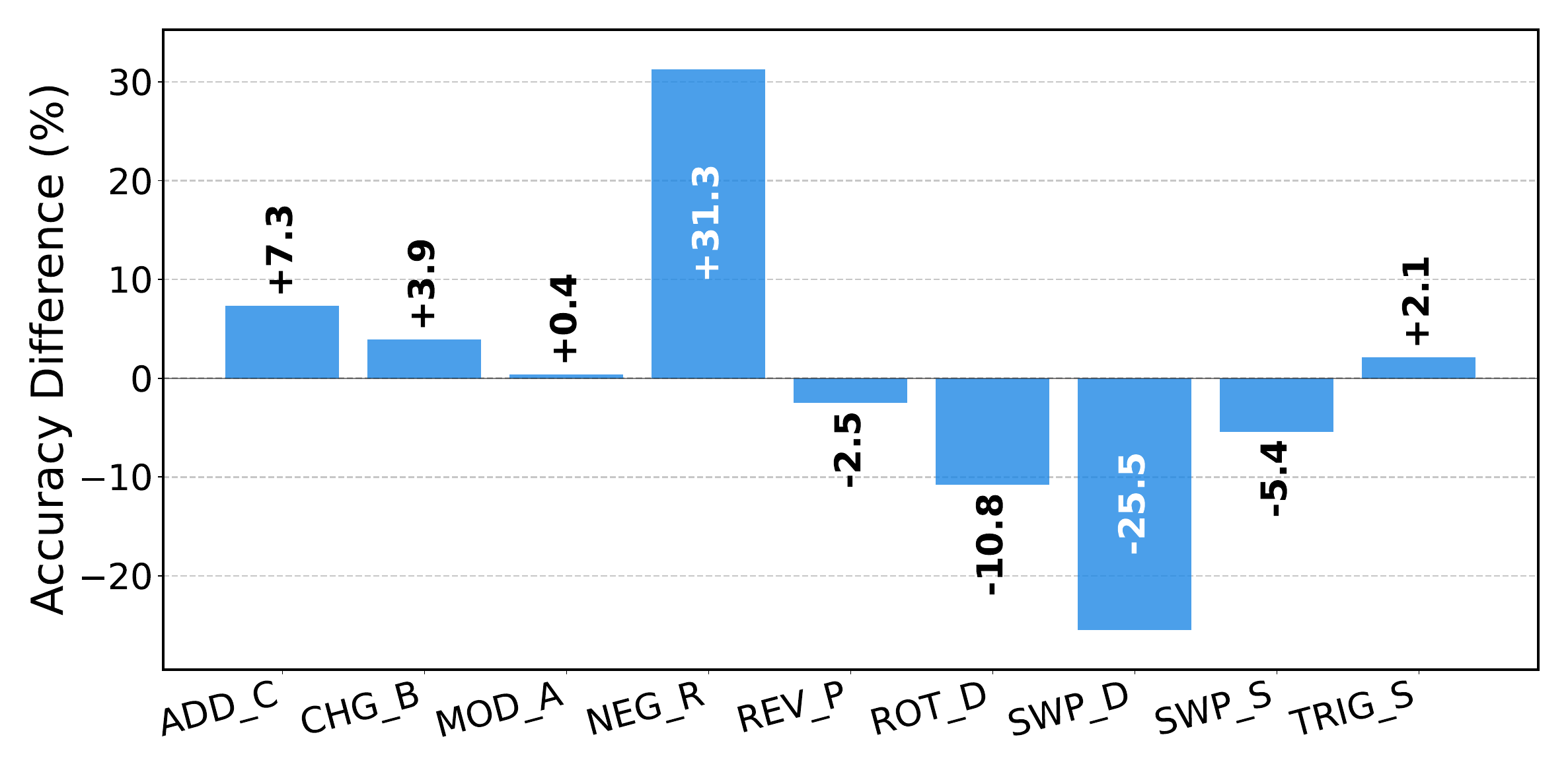}
\vspace{-4ex}
    \caption{Accuracy difference between fine-tuned Qwen2.5 7B and Qwen2.5 7B Instruct. Fine-tuned on \textit{all but one} transformation and tested on the remaining one.}
 \label{fig:train_loo}
\vspace{-2ex}
\end{figure}

 Our first experiment (Figure~\ref{fig:generalized}) shows the model performed well on the data it was trained on, but largely failed to generalize to new, unseen data, implying overfitting. On the training set, accuracy improved across most tasks, with gains of up to +60.6\% for \texttt{TRIG\_S}. On the test set, however, these improvements were much smaller, and for some transformations, performance even worsened, declining by $-21.4\%$ for \texttt{SWP\_D} and $-14.3\%$ for \texttt{ROT\_D}. The one clear exception was \texttt{NEG\_R}, where test set improvement (+55.1\%) exceeded the training set gain (+46.8\%), suggesting the model successfully generalized this simple, global rule but not the more complex transformations. Notably, these results come from the training configuration that performed best on a separate validation set, suggesting that other training approaches would not lead to better generalization.

The remaining experiments confirmed this generalization failure. Specializing in a single transformation (Figure~\ref{fig:train_one}) hurt performance on all other tasks, showing no skill transfer. Likewise, in a leave-one-out test (Figure~\ref{fig:train_loo}), the model failed to generalize to the held-out task with performance often degrading (e.g., -25.5\% for \texttt{SWP\_D}) or showing marginal gains (e.g., +0.4\% for \texttt{MOD\_A}). The one clear exception was again \texttt{NEG\_R}, which improved by +31.3\%. This indicates that the model does not develop a general inductive learning skill from its training, succeeding on a novel task only if the rule is exceptionally simple.

These experiments show that, in our setting, supervised fine-tuning does not impart a general skill of inducing arithmetic rules; instead, the model memorizes superficial patterns from training.

\section{Conclusion}
We introduced a novel framework, \mathemagic{}, that uses procedurally generated, counterfactual arithmetic to test reasoning while preventing data contamination. At test time, test replication and variations are controlled by a random seed. 
We conducted seven analyses with careful control and offer two key findings: models are strong at applying given arithmetic rules but struggle to infer them from examples, and this inductive skill is not easily taught through supervised fine-tuning. This is corroborated by our further fine-tuning experiments, where models are seen to memorize superficial patterns but not generalize inductive reasoning better.

\section*{Limitations}
We did not establish a human performance upper bound on the benchmark due to both the cost and the difficulty in defining a comparable time constraint for humans (considering the challenge of a human annotator reading 512 math examples). Our benchmark covers arithmetic expressions only, mostly with a single operation, so our findings may not extend to algebra, multi-step word problems or proofs. Additionally, we only tested our approach using mathematics and thus do not know how it extends to formal grammars in other domains, for example propositional logic or inferring grammatical rules for machine translation.

\section*{Ethical Considerations}
We contribute to the topic of data contamination and help the community to assess research progress more genuinely and fairly. All model and data artifacts were used in full compliance with their respective licenses.

\section*{Acknowledgments}

\lettrine[image=true, lines=2, findent=1ex, nindent=0ex, loversize=.15]{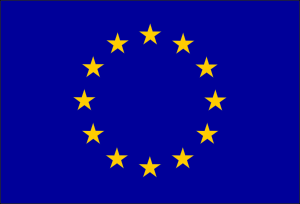}%
This project has received funding from UK Research and Innovation (UKRI) under the UK government’s Horizon Europe funding guarantee [grant number 10052546]. 

Dayyán O'Brien is also funded by a G-Research NextGen Scholarship, part of the UKRI AI Centre for Doctoral Training in Responsible and Trustworthy in-the-world Natural Language Processing (grant ref: EP/Y030656/1). 

We thank Alison Chi, Chenxin Diao, Sherrie Shen, Nataliya Stepanova, and Yijun Yang for helpful suggestions.

\bibliography{custom}

\appendix

\newpage
\section{Prompts}
\label{sec:prompts}

\subsection{System Rule}
A single, fixed system rule was used for all experiments to prime the models for the task. This rule defines the persona of the AI assistant and the expected numerical format for answers.

\begin{promptbox}{System Rule}
You are a helpful AI assistant in a world where mathematical notation has changed. Use only the provided context. Always answer using decimal numbers only.
\end{promptbox}

\subsection{Verifier}
This prompt is used to generate Python functions. It is given, as examples, a set of different transformations and its code in addition to the reasoning provided by the model that performed the induction.

\begin{promptbox}
SYSTEM
You are an expert programmer specialising in code generation. Your task is to analyse a natural language reasoning chain that describes a mathematical transformation and convert it into a single, executable Python function named `solver`.

USER
Your function must adhere to these rules:
1.  The function signature must be `def solver(problem):`.
2.  The function should accept a string `problem` (e.g., '52 - 8') as input.
3.  The output must be **only** the Python code for the function, with no surrounding text, explanations, or markdown fences.
4.  If trigonometric functions or other advanced maths is needed, you must `import math`. For regular expressions, `import re`.
5.  Use `eval()` to perform the standard arithmetic operations within the problem string.
6.  If the reasoning output is broken or does not have a clear function, output "UNKNOWN FUNCTION".

Here are examples of reasoning chains and their corresponding correct `solver` functions.

---
**EXAMPLE 1:**

**REASONING:**
1. Interpret the transformation rules: The problem is a simple subtraction operation where we need to subtract 8 from 52.
2. Apply them step by step:
   - Start with the number 52.
   - Subtract 8 from 52.
3. Justify each transformation:
   - The operation is straightforward: 52 minus 8 equals 44. This is a basic arithmetic operation.
4. Final Answer: \boxed{44}.

**CODE:**
def solver(problem):
    return eval(problem)
---
....
---
**QUESTION**

**REASONING:**
{example['output_raw']}

**CODE:**
\end{promptbox}

\subsection{Instruction Variants}
Four distinct instruction variants were tested to assess the impact of output formatting requirements on model performance. Each prompt included exactly one of these variants appended to the System Rule.

\begin{promptbox}{Instruction Variant 1}
Instruction: \
Present your solution in the following format:
1. Interpret the transformation rules.
2. Apply them step by step.
3. Justify each transformation.
4. Final Answer: LaTeX  \boxed\{answer\}.
\end{promptbox}

\begin{promptbox}{Instruction Variant 2}
Instruction: \
Solve the problem by:
1. Explaining the transformation logic.
2. Showing the intermediate steps clearly.
3. Justifying each step briefly.
4. Final Answer: LaTeX  \boxed\{answer\}.
\end{promptbox}

\begin{promptbox}{Instruction Variant 3}
Instruction: \
Solve the problem and present only the final result as LaTeX  \boxed\{answer\}.
\end{promptbox}

\begin{promptbox}{Instruction Variant 4}
Instruction: \
Compute the correct result. Provide the answer only. Output:  \boxed\{answer\} \
\end{promptbox}

\subsection{Experimental Approaches}
Two primary experimental approaches were designed to evaluate different modes of reasoning.

\subsubsection{Descriptive Approach (Hint-Based)}
In this approach, the prompt explicitly provides the model with a natural language hint describing the mathematical transformations. This tests the model's ability to interpret and apply predefined rules.

\paragraph{Generated Hint Examples}
The hint text is generated by the \texttt{transforms\_to\_prompt} function based on the specific transformation applied. Below are examples of hints for each transformation type.

\begin{description}[leftmargin=!,labelwidth=\widthof{\bfseries Modular Arithmetic},itemsep=5pt,font=\footnotesize]
    \item[Swap Symbols] \texttt{\footnotesize The operator mapping was applied, where operator '+' is transformed into '*' and operator '/' is transformed into '-'.}
    \item[Swap Digits] \texttt{\footnotesize The digits 2 and 8 were exchanged.}
    \item[Reverse Place Values] \texttt{\footnotesize The place values of digits were reversed (e.g., 123 $\rightarrow$ 321).}
    \item[Rotate Digits] \texttt{\footnotesize Each number had its digits rotated right by 2 position(s), excluding decimal parts.}
    \item[Add Constant] \texttt{\footnotesize A constant value of 5 was added to the expression.}
    \item[Negate Result] \texttt{\footnotesize The entire expression was negated.}
    \item[Modular Arithmetic] \texttt{\footnotesize Modulo arithmetic was applied with modulo 7.}
    \item[Trig Substitution] \texttt{\footnotesize Trigonometric substitution was applied: replaced 45 with math.cos(90°).}
    \item[Change Base] \texttt{\footnotesize The numeral system was changed to base 8.}
\end{description}

Additionally, a concluding note is often appended, such as: \texttt{Note: Digit transformations modify the representation of numbers, including the final computed result.}

\subsubsection{Inductive Approach (Example-Based)}
In this approach, the prompt contains no hint. The model must infer the hidden mathematical rules solely from the provided few-shot examples (or from the problem statement in the zero-shot case). This tests the model's ability to recognize patterns and generalize from data.

\subsection{Complete Prompt Examples}

The following examples illustrate the raw prompt structure for different experimental configurations, using a hypothetical transformation where the digits `1` and `7` are swapped.

\subsubsection{Inductive, 2-Shot, Instruction Variant 1}
The model must infer the ``swap 1 and 7'' rule from the two examples and provide a step-by-step solution.

\begin{promptbox}{System Message}
You are a helpful AI assistant in a world where mathematical notation has changed. Use only the provided context. Always answer using decimal numbers only. Instruction:
Present your solution in the following format:
1. Interpret the transformation rules.
2. Apply them step by step.
3. Justify each transformation.
4. Final Answer: LaTeX  \boxed\{answer\}.
\end{promptbox}

\begin{promptbox}{User Message}
Examples:
1. 10 * 3 =  \boxed\{210\}
2. 81 - 1 =  \boxed\{80\}

Problem:
11 + 77 =  \boxed\{?\}
\end{promptbox}

\subsubsection{Descriptive, 0-Shot, Instruction Variant 4}
The model is explicitly told the rule via a hint and must provide only the final answer.

\begin{promptbox}{System Message}
You are a helpful AI assistant in a world where mathematical notation has changed. Use only the provided context. Always answer using decimal numbers only.

Instruction:
Compute the correct result. Provide the answer only. Output:  \boxed\{answer\}
\end{promptbox}

\begin{promptbox}{User Message}
Problem:
(Hint: The digits 1 and 7 were exchanged.)
11 + 77 =  \boxed\{?\}
\end{promptbox}

\section{Hyperparameters}
The following table details the hyperparameters used for all fine-tuning experiments. The hyperparameters were selected through manual tuning. We chose the final values based on the best performance observed on a validation set.

\begin{table}[h!]
\centering\small
\caption{Key hyperparameters for QLoRA fine-tuning.}
\label{tab:hyperparameters}
\begin{tabular}{ll}
\toprule
\textbf{Hyperparameter} & \textbf{Value} \\
\midrule
\multicolumn{2}{l}{{Model \& Architecture}} \\
\quad Base Model & \texttt{Qwen/Qwen2.5-7B} \\
\quad Fine-tuning Method & QLoRA \\
\quad LoRA Rank ($r$) & 64 \\
\quad LoRA Alpha ($\alpha$) & 128 \\
\addlinespace
\multicolumn{2}{l}{{Training \& Optimization}} \\
\quad Learning Rate & $5 \times 10^{-4}$ \\
\quad LR Scheduler & Constant \\
\quad Warmup Ratio & 0.03 \\
\quad Weight Decay & 0.01 \\
\quad Effective Batch Size & 64 \\
\quad Training Epochs & 5 \\
\quad Max Sequence Length & 512 \\
\quad Precision & bf16 \\
\quad Seed & 8 \\
\bottomrule
\end{tabular}
\end{table}

\subsection{Computational Resources}
Our experiments were conducted on a local computing infrastructure and via commercial APIs. Our local hardware consists of 8x NVIDIA RTX 3090 GPUs. The fine-tuning experiments for the Qwen2.5 7B were performed on this hardware, with each run taking approximately 30 minutes on 4 NVIDIA RTX 3090 GPUs. All inference jobs for the open-source models (Llama3, Qwen2.5) were also run on our local infrastructure. While the total time is difficult to aggregate, individual evaluation jobs typically took 5-10 minutes to complete. For experiments involving GPT-4o-mini, GPT-5-nano, and GPT-5, we used the official OpenAI API. For Gemini, we used the official Google API. A precise token count is not reported, as we were unable to access the detailed historical usage data for the full set of experiments. 

\section{Licenses}
All code for our benchmark's dataset generation and evaluation is publicly released at \url{https://github.com/dayyanobrien/mathemagic}.

The base mathematical expressions for our benchmark were extracted from GSM8K, which is available under the MIT license. The Qwen2.5 Instruct \citep{qwen2025qwen25technicalreport} series and Llama3 8B Instruct \citep{grattafiori2024Llama3herdmodels} were used in accordance with their respective community licenses. GPT models \citep{openai2024gpt4omini, openai2025gpt5} were accessed via the official OpenAI API, and our usage complied with its terms of service for research purposes.

The data used and generated in this study consists exclusively of mathematical expressions (e.g., "15+4=19") and their symbolic, counterfactual transformations. By its nature, this data format does not contain any personally identifiable information or any other sensitive information.

\section{Expanded Results}

\subsection{Fine-tuning} 
\begin{figure}[!ht]
\includegraphics[width=\columnwidth]{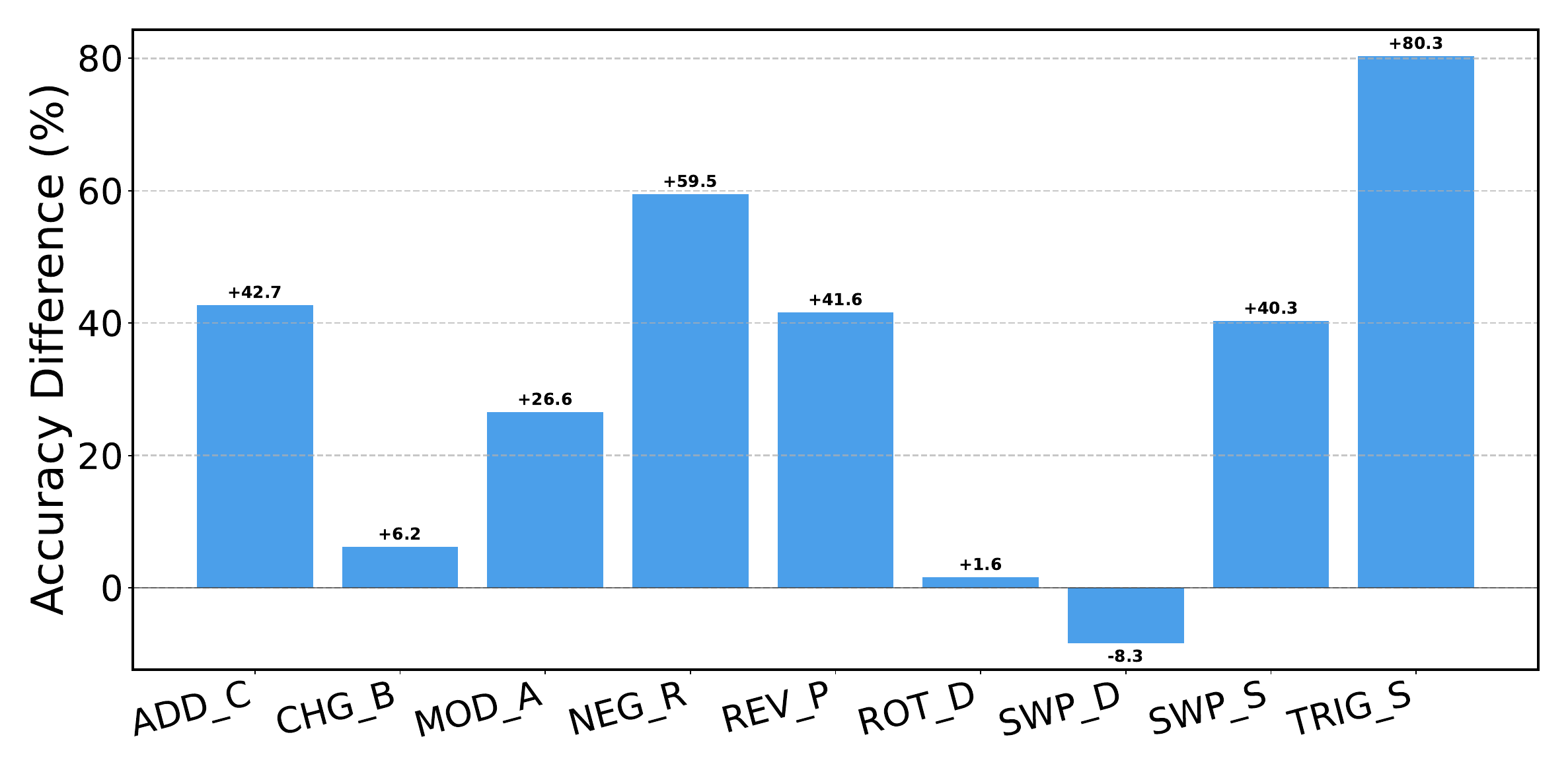}
\caption{Accuracy difference between a model fine-tuned on a single transformation and the baseline model, when tested on that same transformation. Each bar represents a model trained exclusively on one transformation type. The model is fine-tuned from the base Qwen2.5 7B; the y-axis shows the percentage point change in accuracy compared to Qwen2.5 7B Instruct.}
\label{fig:train_incl_incl}
\end{figure}

\paragraph{If you fine-tune on one transform, can it be learned?}
To determine if a model can learn a specific transformation rule through fine-tuning, we trained a separate model for each of our nine transformations. Each model was fine-tuned exclusively on examples of a single transformation type (e.g., one model for \texttt{ADD\_C}, another for \texttt{CHG\_B}, etc.). We then evaluated each specialized model on a test set corresponding to the transformation it was trained on. Figure~\ref{fig:train_incl_incl} shows that this targeted fine-tuning generally leads to substantial performance gains. Models fine-tuned on \texttt{TRIG\_S} and \texttt{NEG\_R}, for example, saw accuracy improvements of over 50 percentage points compared to the baseline. However, the success is not uniform. fine-tuning on more complex procedural tasks like \texttt{CHG\_B} and \texttt{ROT\_D} yielded only modest improvements, while the model trained on \texttt{SWP\_D} actually performed worse. This indicates that while models can be taught to master specific rules, their ability to learn is highly dependent on the nature of the transformation.

\subsection{Variants}
\label{app:variants}

\begin{figure*}[p]
  \centering
  \includegraphics[width=\textwidth]{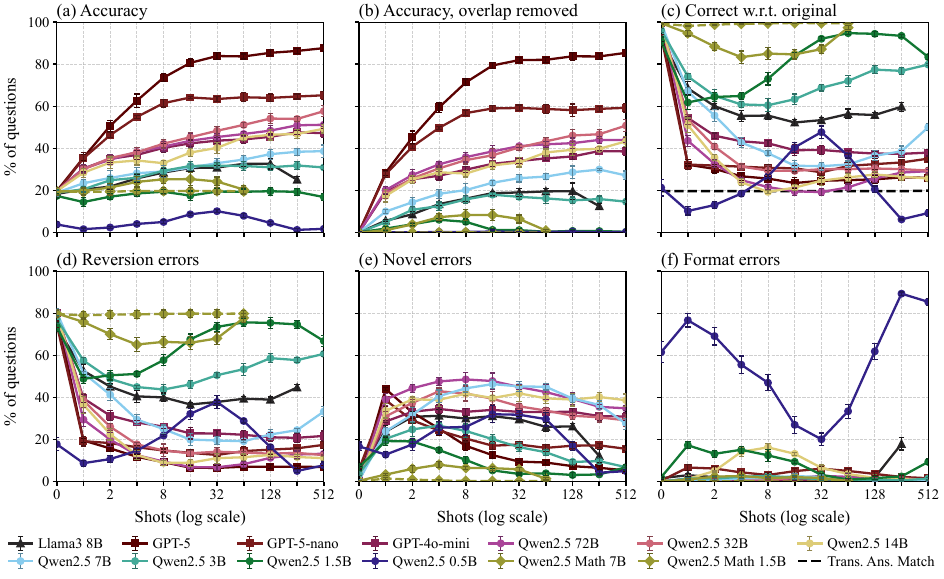}
  \caption{Results with variant 1 (inductive setting, mean over the five seeds). (a)~Accuracy under the transformation. (b)~Accuracy after removing coincidentally correct questions. (c)~Answers correct under ordinary arithmetic. (d)~Reversion errors: answering with ordinary arithmetic when it is wrong. (e)~Novel errors: wrong answers that match neither. (f)~Format errors: the answer cannot be parsed. Llama3 8B is evaluated up to 256 shots and the Qwen2.5-Math models up to 64.}
  \label{fig:variant1}

  \vspace{1.5em}
  \includegraphics[width=\textwidth]{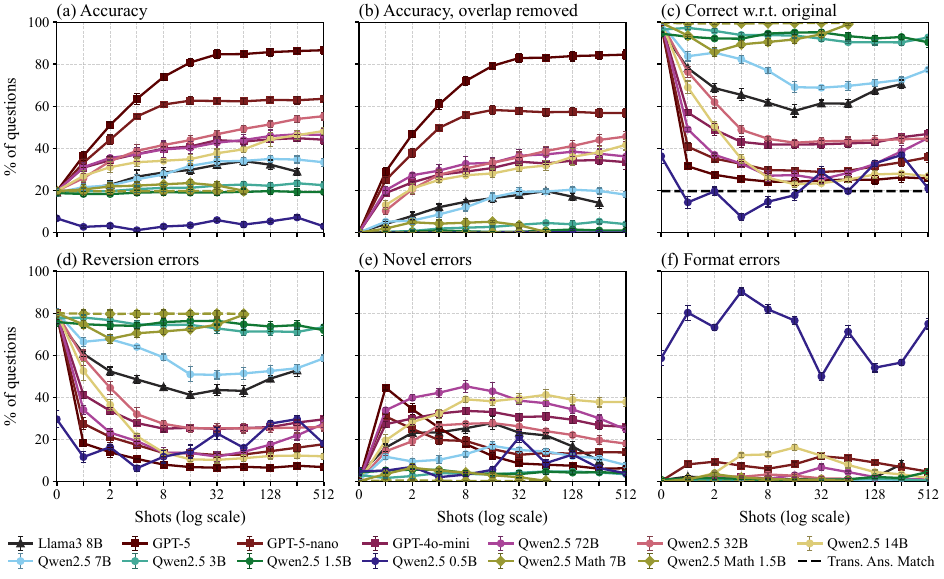}
  \caption{Results with variant 2 (inductive setting, mean over the five seeds). Panels as in Figure~\ref{fig:variant1}.}
  \label{fig:variant2}
\end{figure*}

\begin{figure*}[p]
  \centering
  \includegraphics[width=\textwidth]{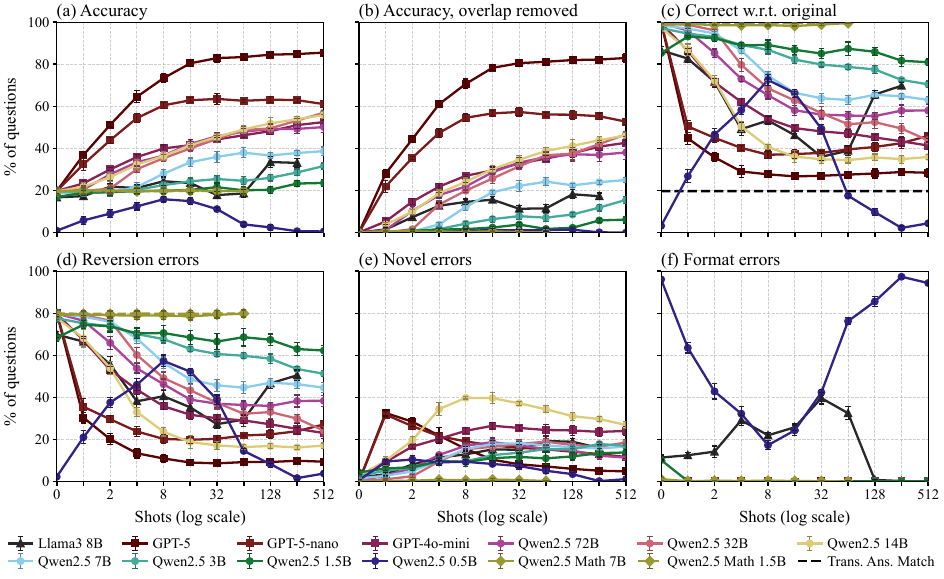}
  \caption{Results with variant 3 (inductive setting, mean over the five seeds). Panels as in Figure~\ref{fig:variant1}.}
  \label{fig:variant3}

  \vspace{1.5em}
  \includegraphics[width=\textwidth]{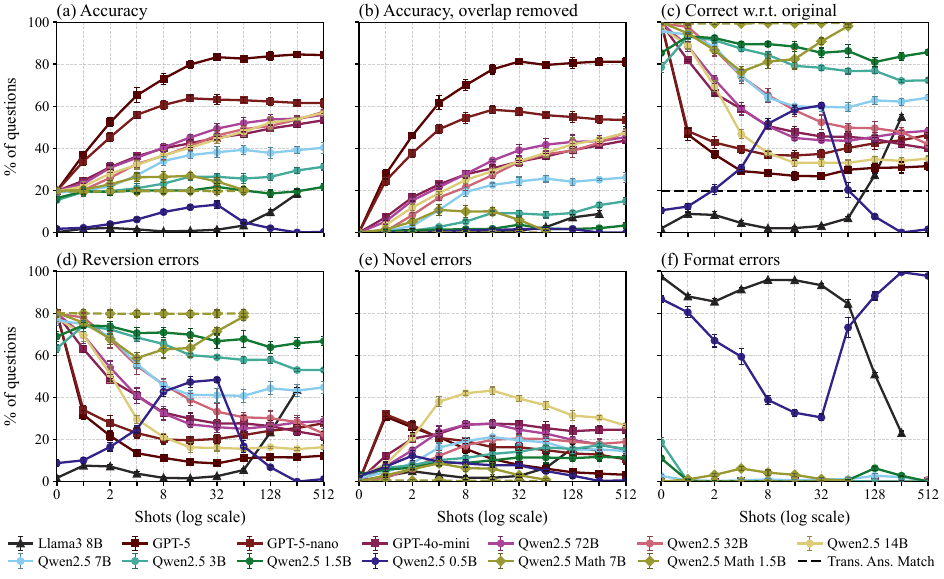}
  \caption{Results with variant 4 (inductive setting, mean over the five seeds). Panels as in Figure~\ref{fig:variant1}.}
  \label{fig:variant4}
\end{figure*}

This section presents detailed results across all four prompt instruction variants (Appendix~\ref{sec:prompts}). Each variant phrases and formats the task instructions differently. Figures~\ref{fig:variant1}--\ref{fig:variant4} show six metrics for each variant: accuracy, accuracy with overlapping questions removed, answers that are correct under ordinary arithmetic, reversion errors, novel errors and format errors. Together they show how consistent model performance is across instruction formulations.

\paragraph{Overlapping questions.} For some questions, the transformed answer happens to equal the ordinary-arithmetic answer (400 of the 2,025 questions across the five seeds, 19.8\%). On these, a model that ignores the rule and does ordinary arithmetic is scored as correct. Panel~(b) of each figure reports accuracy with these
questions removed. Zero-shot accuracy then falls to at most 0.4\% for every model, and the gain from in-context examples is larger: with variant~1, Qwen2.5 72B goes from 0.0\% at zero shots to 44.1\% at 512 shots, compared with 19.8\% and 51.3\% on all questions.

\renewcommand{\dblfloatpagefraction}{0.5}

\end{document}